\def\year{2022}\relax
\documentclass[letterpaper]{article} 
\usepackage{aaai22}  
\usepackage{times}  
\usepackage{helvet}  
\usepackage{courier}  
\usepackage[hyphens]{url}  
\usepackage{graphicx} 
\usepackage{natbib}  
\usepackage{caption} 
\DeclareCaptionStyle{ruled}{labelfont=normalfont,labelsep=colon,strut=off} 
\usepackage{algorithm}
\usepackage{algorithmic}
\usepackage{bm}
\usepackage{booktabs}
\usepackage{amsmath}
\usepackage{amsfonts}	
\usepackage{multirow}
\usepackage{newfloat}
\usepackage{listings}
\floatstyle{ruled}
\newfloat{listing}{tb}{lst}{}
\floatname{listing}{Listing}
\title{UCTransNet: Rethinking the Skip Connections in U-Net from a Channel-wise Perspective with Transformer}
\author {
	Haonan Wang\textsuperscript{\rm 1,\rm 2},
	Peng Cao\textsuperscript{\rm 1,\rm 2}\footnote{Corresponding author.},
	Jiaqi Wang\textsuperscript{\rm 1,\rm 2},
	Osmar R. Zaiane\textsuperscript{\rm 3}
}
\affiliations {
	\textsuperscript{\rm 1} Computer Science and Engineering, Northeastern University, Shenyang, China
	\\
	\textsuperscript{\rm 2} Key Laboratory of Intelligent Computing in Medical Image of Ministry of Education, Northeastern University, China
	\\
	\textsuperscript{\rm 3} Amii, University of Alberta, Edmonton, Canada
	\\
	haonan1wang@gmail.com, caopeng@mail.neu.edu.com, wjq010222@gmail.com, zaiane@cs.ualberta.ca
}

\usepackage{bibentry}

\begin{document}
	
	\maketitle
	\begin{abstract}
		Most recent semantic segmentation methods adopt a U-Net framework with an encoder-decoder architecture. It is still challenging for U-Net with a simple skip connection scheme to model the global multi-scale context: 1) Not each skip connection setting is effective due to the issue of incompatible feature sets of encoder and decoder stage, even some skip connection negatively influence the segmentation performance; 2) The original U-Net is worse than the one without any skip connection on some datasets.
		Based on our findings, we propose a new segmentation framework, named UCTransNet (with a proposed CTrans module in U-Net), from the channel perspective with  attention mechanism. 
		Specifically, the CTrans (Channel Transformer) module is an alternate of the U-Net skip connections, which consists of a sub-module to conduct the multi-scale \textbf{C}hannel  \textbf{C}ross fusion with \textbf{T}ransformer  (named CCT) and a sub-module \textbf{C}hannel-wise \textbf{C}ross-\textbf{A}ttention (named CCA) to guide the fused multi-scale channel-wise information to effectively connect to the  decoder features for eliminating the ambiguity.  
		Hence, the proposed connection consisting of the CCT and CCA  is able to replace the original skip connection to solve the semantic gaps for an accurate automatic medical image segmentation.
		The experimental results suggest that our UCTransNet produces more precise segmentation performance and achieves consistent improvements over the state-of-the-art for semantic segmentation across different datasets and conventional architectures involving transformer or U-shaped framework.
		Code: \url{https://github.com/McGregorWwww/UCTransNet}.

	\end{abstract}

	\section{Introduction}
	
	
	Medical imaging is considered as a vital technique to assist doctors to evaluate disease and to optimise prevention and control measures. Segmentation and the subsequent quantitative assessment of target object in medical images provide valuable information for the analysis of pathologies and are important for planning of treatment strategies, monitoring of disease progression and prediction of patient outcome.
	Recent approaches \cite{FullyConvolutionalNetworks_2015,PyramidAttentionAggregation_2020,SegmentingMedicalMRI_2020} to semantic segmentation typically rely on convolutional encoder-decoder architectures where the encoder generates low-resolution image features and the decoder up-samples features to segmentation maps with per pixel class scores.
	U-Net \cite{UNetConvolutionalNetworks_2015} is the most widely used encoder-decoder network architecture for medical image segmentation, since the encoder captures the low-level and high-level features, and the decoder combines the semantic features to construct the final result. 
	The skip connection can help propagate the spatial information that gets lost during the pooling operation to help recover the full spatial resolution through the encoding-decoding process.
	To investigate it, we conduct an in-depth study of U-Net and observe several major limitations according to our analysis on multiple datasets. We find that it is still challenging for U-Net with a simple skip connection scheme to model the global multi-scale context for assisting the decoding process without considering the semantic gap. 
	\begin{figure}[t]
		\centering
		\includegraphics[width=1.\columnwidth]{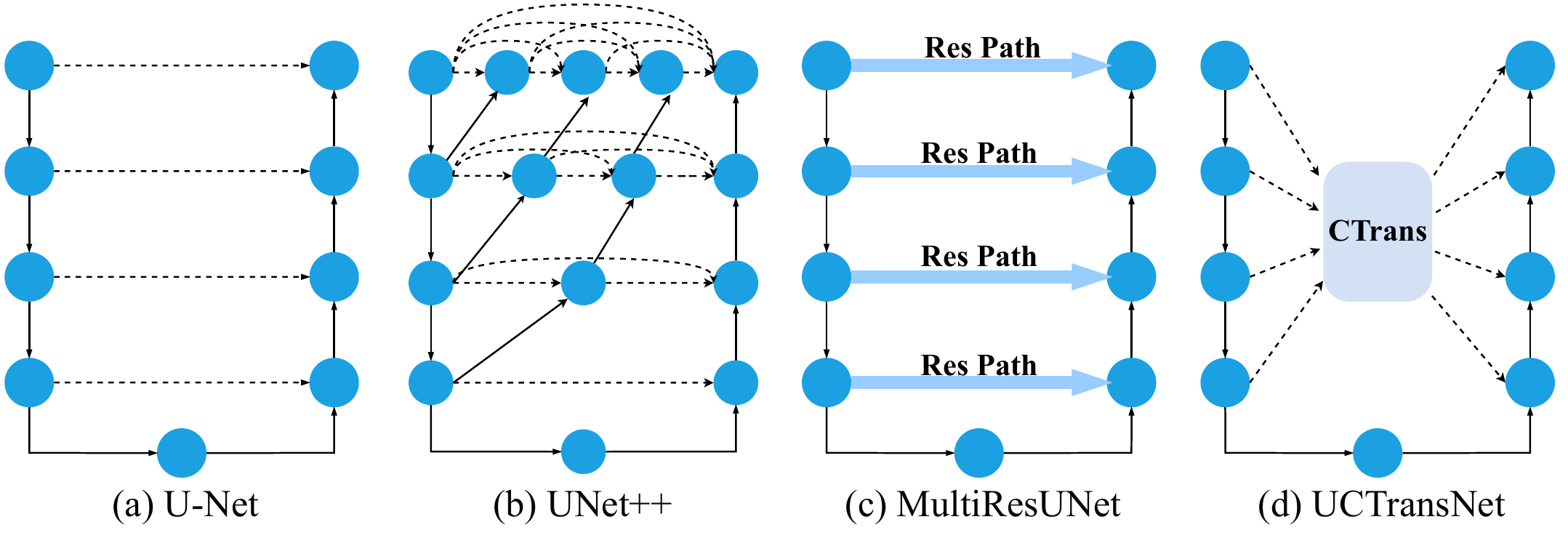} 
		\caption{Comparison of the skip connection scheme among the proposed UCTransNet (d) and other models. The dash lines denote the skip connections.}
		\label{SkipCompare}
	\end{figure}
	\begin{figure*}[!ht]
		\centering
		\includegraphics[width=0.87\textwidth]{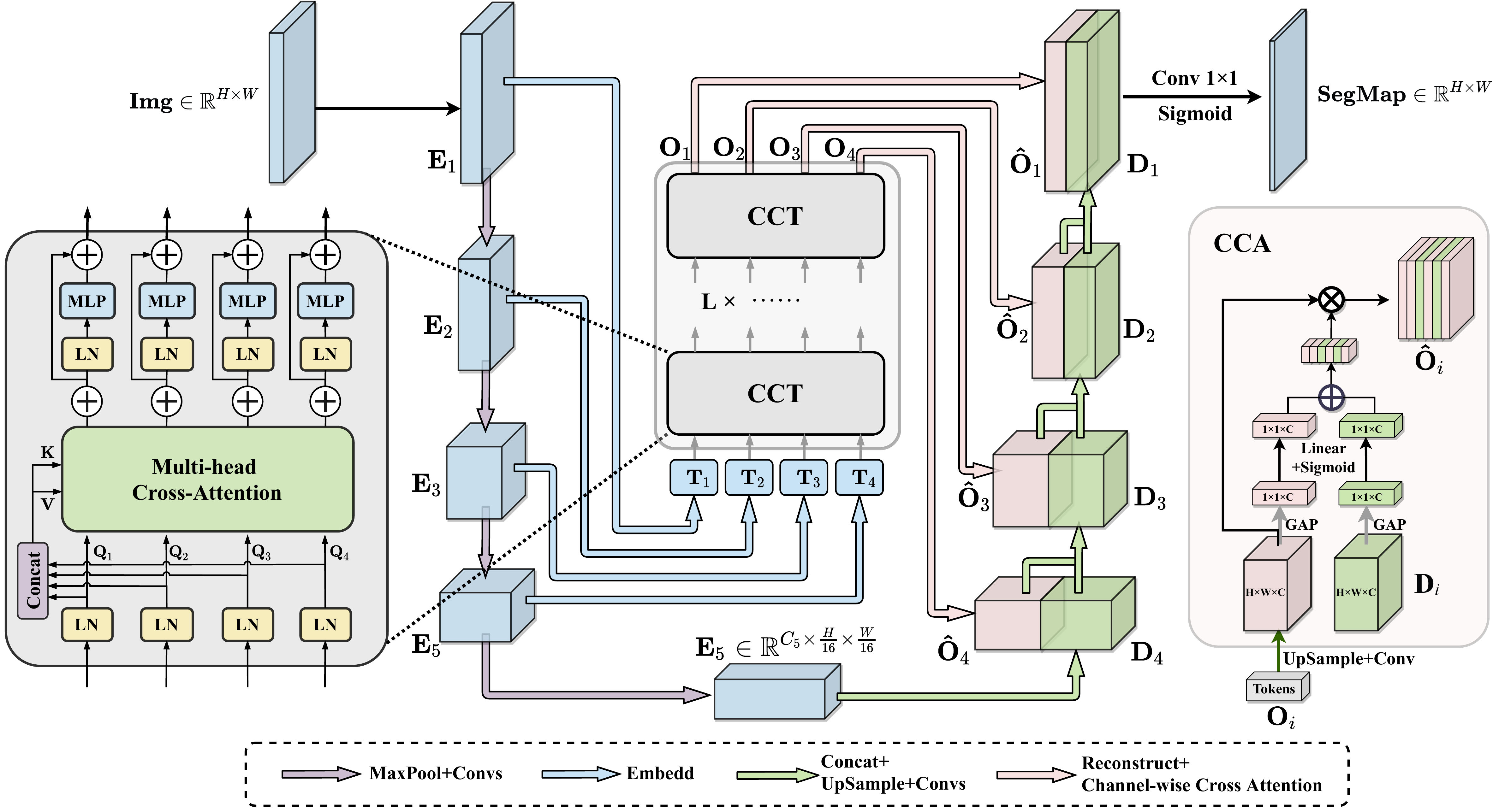} 
		\caption{Illustration of the proposed UCTransNet. We replace the original skip connections by CTrans consisting of two  components: Channel-wise Cross fusion Transformer (CCT) and Channel-wise Cross Attention (CCA).}
		\label{Framework}
	\end{figure*}
	It is necessary to find an effective way to fuse features for precise medical image segmentation.
	There essentially are two key issues for the extension of U-Net: which layers of the features in the encoders are connected to the decoders for modeling a global contexts through aggregating multi-scale features, and how to effectively fuse the features with possible semantic gap instead of simply concatenating?
	There exist two semantic gaps: semantic gap among the multi-scale encoder features and between the stages of the encoder and decoder, limiting the segmentation performance.
	To overcome this aforementioned limitation, a number of approaches have been introduced recently to alleviate the discrepancy when fusing these two incompatible sets of features.
	One approach is to directly replace the plain skip connections with the nested dense skip pathways for medical image segmentation.
	The most representative method is UNet++ \cite{UNetNestedUNet_2018} which narrows the semantic gap between the encoder and decoder sub-networks by introducing dense connectivity with a series of convolutions and achieves better segmentation performance. It is an improvement over the restrictive skip connections in U-Net requiring the fusion of only same-scale feature maps.
	The other approach focuses on strengthening the skip connections by introducing additional non-linear transformations on the features propagating from the encoder stage, which should account for or somewhat balance the possible semantic gaps \cite{MultiResUNetRethinkingUNet_2020}.

	Despite achieving good performance, both works above are still incapable of effectively exploring sufficient information from full scales.
	Capturing multi-scale features is essential for resolving complex scale variations in medical image segmentation. 
	Driven by the important issues, a question arises: how to sufficiently  bridge the  semantic gap between the encoder and decoder  through multi-scale channel-wise  information fusion by  effectively capturing the non-local semantic dependencies.
	In this paper, we rethink the skip connection design and propose an alternative method for better connecting the features between the encoder and decoder stages.
	Different channels usually focus on different semantic patterns, adaptively fusing sufficient channel-wise features is favorable for the complex medical image segmentation.
	To this end, we propose an end-to-end deep learning network called UCTransNet, which takes U-Net as the main structure of the network. More specifically, we firstly propose a  Channel-wise Cross Fusion Transformer (CCT) to fuse the multi-scale context with cross attention from the channel-wise perspective. It aims at capturing local cross-channel interaction to achieve an adaptive scheme for effectively fusing the multi-scale channel-wise features with possible scale semantic gap through collaborative learning rather than independent connection. 
	On the other hand, we propose another channel-wise cross attention (CCA) module for fusing the fused multi-scale features and the features from decoder stages to solve the inconsistent semantic level.  
	Both cross attention  modules are called CTrans (Channel Transformer), which can establishes the association between encoder and decoder by exploring the multi-scale global context and replace the original skip connections to solve the semantic gaps for improved segmentation performances.
	Both proposed modules can be easily embedded in and applied for the U-shape networks in medical image segmentation tasks.
	Extensive experiments show that UCTransNet can greatly improve conventional segmentation pipelines by the following absolute gains of 4.05\% Dice,  7.98\% Dice  and 9.00\% Dice  over U-Net on GlaS, MoNuSeg and Synapse datasets, respectively. 
	Moreover, we made a thorough analysis to investigate how the feature interactions work. Besides, previous works have combined both Transformers and U-Net to explicitly model long-range spatial dependency \cite{TransUNetTransformersMake_2021,TransFuseFusingTransformers_2021}. The results demonstrate that channel-wise fusion transformer scheme generally leads to a better performance than the methods incorporating the transformer to
	replace the convolution operation.
	We argue that UCTransNet can serve as strong skip connection scheme for medical image segmentation.
	
	Our contributions are three-fold.
	\textbf{1)} Our study is the first work that sufficiently explores the potential weakness of skip connections in U-Net on multiple datasets and finds that the independent simple copying is not appropriate. 
	\textbf{2)} We suggest a new perspective to boost semantic segmentation performance, i.e. bridging the semantic and resolution gap between low-level and high-level features by a more effective feature fusion with multi-scale channel-wise cross attention for capturing more sophisticated channel-wise dependencies.
	\textbf{3)} Our method is a more appropriate combination of U-Net and Transformer with less computational cost and higher performance. In comparison to other state-of-the-art segmentation methods, the experimental results present better performances on all the three public datasets.


	

	
	\section{Related Works}

	\subsection{Transformers for Medical Image Segmentation}
	Recently, Vision Transformer (ViT) \cite{ImageWorth16x16_2020} achieved state-of-the-art on ImageNet classification by directly applying Transformers with global self-attention to full-sized images.
	Due to the success of Transformers in many computer vision fields, a new paradigm for medical image segmentation has recently evolved\cite{RethinkingSemanticSegmentation_2020,MultiBranchHybridTransformer_2021, InstancebasedVisionTransformer_2021, MultiCompoundTransformerAccurate_2021, UTNetHybridTransformer_2021,TransFuseFusingTransformers_2021,UNETRTransformers3D_2021}.
	TransUNet \cite{TransUNetTransformersMake_2021} is the first Transformer-based medical image segmentation framework.
	Valanarasu et al. proposed a Gated Axial-Attention model–MedT \cite{MedicalTransformerGated_2021} to overcome the low number of data samples in medical imaging.
	Motivated by Swin Transformer \cite{SwinTransformerHierarchical_2021} which achieved state-of-the-art performance, Swin-Unet \cite{SwinUnetUnetlikePure_2021} proposed the first pure Transformer-based U-shaped architecture which introduced Swin Transformer to replace the convolution blocks in U-Net.
	However, the aforementioned methods mainly focus on the defects of convolution operation rather than the U-Net it-self, thus may cause structural redundancy and prohibitive computational cost.

	\subsection{Skip Connections in U-shaped Nets}
	
	The skip connection mechanism was first proposed in U-Net \cite{UNetConvolutionalNetworks_2015}, which was designed to bridge the semantic gap between encoder and decoder, and have proven to be effective in recovering fine-grained details of the target objects \cite{ImportanceSkipConnections_2016,DeepResidualLearning_2016a, DenselyConnectedConvolutional_2017}.
	Following the popularity of U-Net, many novel models have been proposed such as UNet++ \cite{UNetNestedUNet_2018}, Attention U-Net \cite{AttentionUNetLearning_2018}, DenseUNet \cite{HDenseUNetHybridDensely_2018}, R2U-Net \cite{RecurrentResidualConvolutional_2018a}, and UNet 3+ \cite{UNetFullScaleConnected_2020}, which are specially designed for medical image segmentation and achieve expressive performance.
	Zhou et al. \cite{UNetNestedUNet_2018} believed that the same-scale feature maps from the encoder and decoder networks are semantically dissimilar and thus designed a nested structure named UNet++ which captures multi-scale features to further bridge the gap.
	Attention-UNet proposed cross-attention module which uses coarse-scale features as gating signals to disambiguate irrelevant and noisy responses in skip connections. 
	MultiResUNet \cite{MultiResUNetRethinkingUNet_2020} observed a possible semantic gap between the skipped encoder features and the decoder features in the same level, thus they introduced the Res Path with residual structure to improve the skip connections (see Fig.~\ref{SkipCompare}).
	
	These methods assume that each skip connection has equal contribution, however in the next section we will show that the contributions are different among all the skip connections, some may even harm the final performance.

	\begin{figure}[t]
		\centering
		\includegraphics[width=0.95\columnwidth]{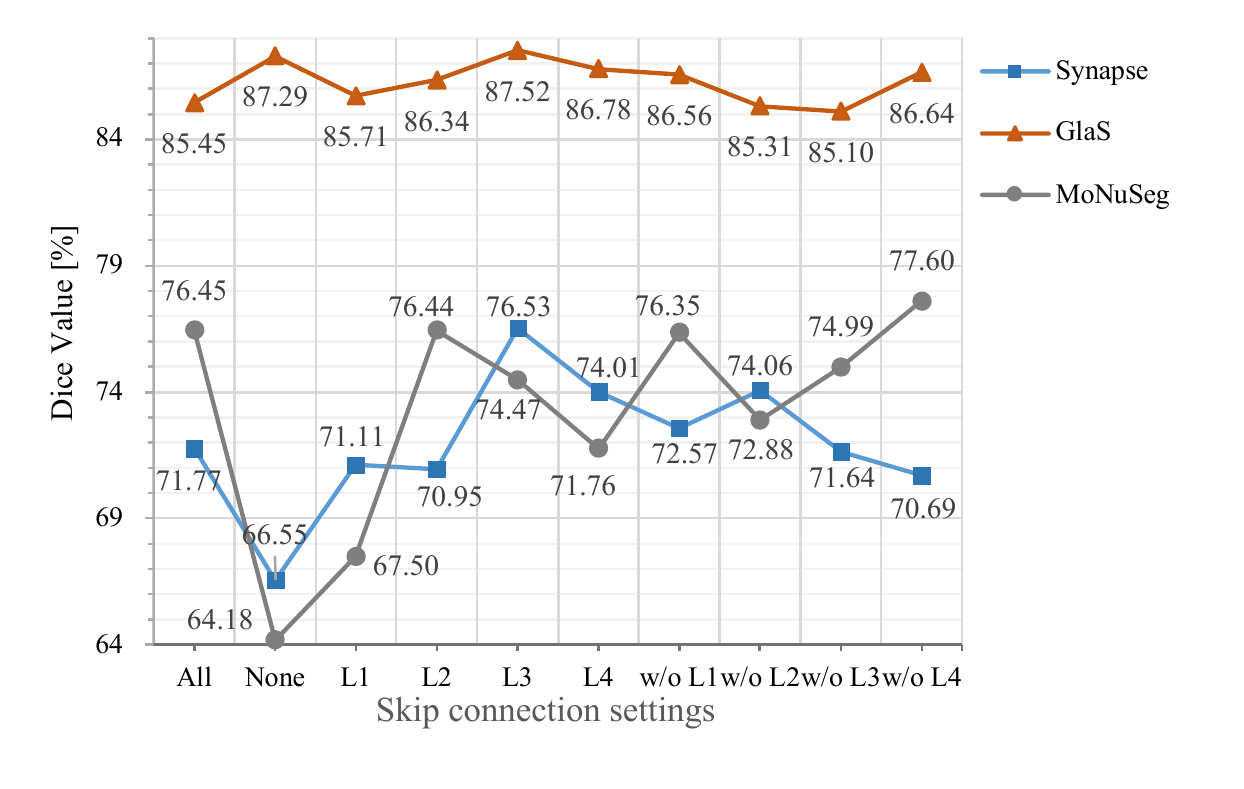} 
		\caption{Analysis of different skip connection layers of U-Net. `All' represents the original U-Net, `L1' represents only the skip connection of level one is kept and `w/o L1' represents only the skip connection of level one is removed.}
		\label{skip}
	\end{figure}
	
	\section{The Analysis of Skip connection } 
	\label{section:analysis}
	
	In this section, we thoroughly analyze the contribution of the skip connection to the segmentation performance on three datasets.  According to the analysis, three findings are highlighted as follows:
	
	\textbf{Finding 1}: The U-net without any skip connection is even better than the original U-Net.
	Comparing the results of Fig.~\ref{skip}, we can find that `U-Net-none' shows the worst performance among the algorithms for almost all metrics on the MoNuSeg dataset. However, `U-Net-none', although without any constraints, still achieves very competitive performance against `U-Net-all' on the GlaS dataset. 
	It demonstrates that the skip connection is not always beneficial for the segmentation.
	
	\textbf{Finding 2}: Although UNet-all performs better than UNet-none, not all skip connections with simple copying are useful for segmentation. The contribution of each skip connection is different.
	We find that the performance range of each skip connection  is [67.5\%,76.44\%] and [52.2\%,62.73\%] with respect to Dice and IOU on  the MoNuSeg dataset. The impact variation is large for the different single skip connection. 
	Furthermore, due to the issue of incompatible feature sets of the encoder and decoder stages, some skip connection negatively influence the segmentation performance. For example,  $L_1$ performs worse than UNet-none in terms of Dice and IOU on the GlaS dataset.
	The result does not demonstrate that many features from the encoder stage are not informative. The reason behind it may be that the simple  copying is not appropriate for the feature fusion.
	
	\textbf{Finding 3}: The optimal combination of skip contributions is different for different datasets, which depends on the scales and appearance of the target lesions.
	We run several ablation experiments to explore the best side output settings. Note that we ignore the combination of two skip connections due to the limited space. As can be seen, the skip connections does not achieve better performance.
	The model w/o $L_4$ is best on the MoNuSeg dataset, whilst to our surprise, the $L_3$  with only one skip connection performs best on the GlaS dataset. 
	These observations suggest that the optimal combination is different for different datasets which further confirms the necessity of introducing more appropriate course of action for the feature fusion rather than simple connection.

	\section{UCTransNet for Medical Image Segmentation}
	
	Fig.~\ref{Framework} illustrates an overview of our UCTransNet framework.
	To the best of our knowledge, current Transformer-based segmentation methods mainly focus on improving the encoder of U-Net, based on its advantage of capturing long-range information. These methods, such as TransUNet \cite{TransUNetTransformersMake_2021} or TransFuse \cite{TransFuseFusingTransformers_2021},  blend the Transformer with U-Net in a simple way, i.e. plugging the Transformer module into the encoder or fusing the both independent branches.
	However, we believe the potential limitation of the current U-Net model is the issue of the skip connection rather than  the encoder of the original U-Net, which is sufficient for the most tasks. 
	As mentioned in the section of skip connection analysis, we observe that the feature from the encoder is inconsistent with that from the decoder, i.e. in some cases, the  shallower layer features with less semantic information  may harm the final performance through the simple skip connection due to the semantic gap between the shallower level encoder and decoder. 
	Inspired by it, we construct the  UCTransNet framework by designing a   channel-wise Transformer module between the vanilla U-Net encoder and decoder to better fuse the encoder features and reduce the semantic gap. Specifically,
	we propose a Channel Transformer (CTrans) to replace the skip connections in U-Net, which consists of two modules: CCT (Channel-wise Cross Fusion Transformer) for the multi-scale encoder feature fusion and CCA (Channel-wise Cross Attention) for the fusion of the decoder features and the enhanced CCT features.
	
	\begin{figure}[t]
		\centering
		\includegraphics[width=0.7\columnwidth]{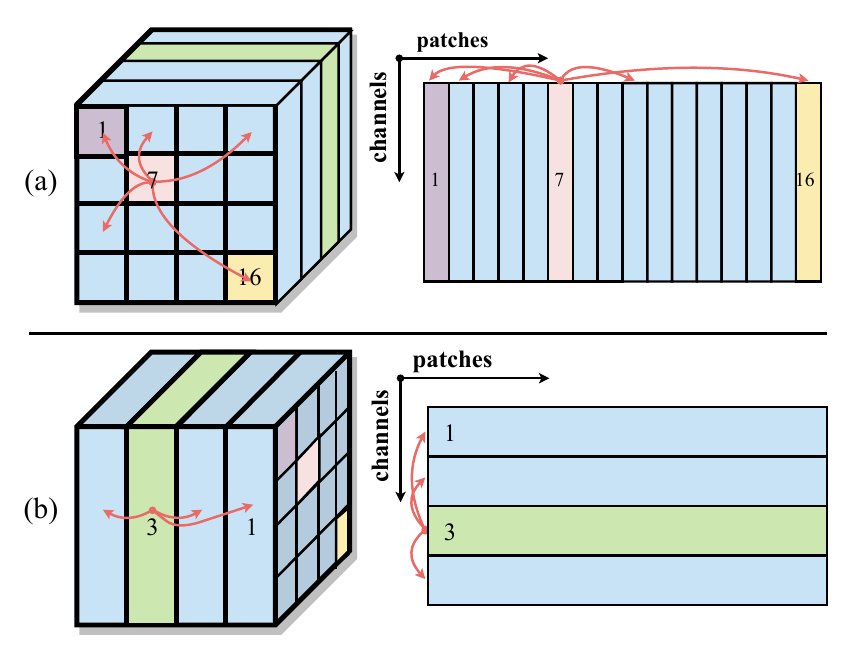} 
		\caption{Comparison between the original self-attention (a) and our proposed channel-wise cross-attention (b).}
		\label{TransComparison}
	\end{figure}

	\subsection{CCT: Channel-wise Cross Fusion Transformer for Encoder Feature Transformation}

	To solve the skip connection issue mentioned before, we propose a new Channel-wise Cross Fusion Transformer (CCT) to fuse the  multi-scale encoder features with the advantage of the long dependency modeling in Transformer.
	The CCT module consists of three steps: multi-scale feature embedding, multi-head channel-wise cross attention and Multi-Layer Perceptron (MLP).  
	
	\subsubsection{Multi-scale Feature Embedding}
	Given the outputs of four skip connection layers $\mathbf{E}_i \in \mathbb{R}^{\frac{HW}{i^2}\times C_i},(i=1,2,3,4)$, we first perform tokenization by reshaping the features into sequences of flattened 2D patches with patch sizes $P,\frac{P}{2},\frac{P}{4},\frac{P}{8}$ respectively, so that the patches can be mapped to the same areas of the encoder features in four scales. We keep the original channel dimensions through this process.
	Then, we concatenate the tokens of four layers $\mathbf{T}_i(i=1,2,3,4), \mathbf{T}_i \in \mathbb{R}^{\frac{HW}{i^2}\times C_i}$ as the key and value $\mathbf{T}_{\Sigma} = \mathrm{Concat}(\mathbf{T}_1,\mathbf{T}_2,\mathbf{T}_3,\mathbf{T}_4)$.

	\begin{figure}[t]
		\centering
		\includegraphics[width=0.9\columnwidth]{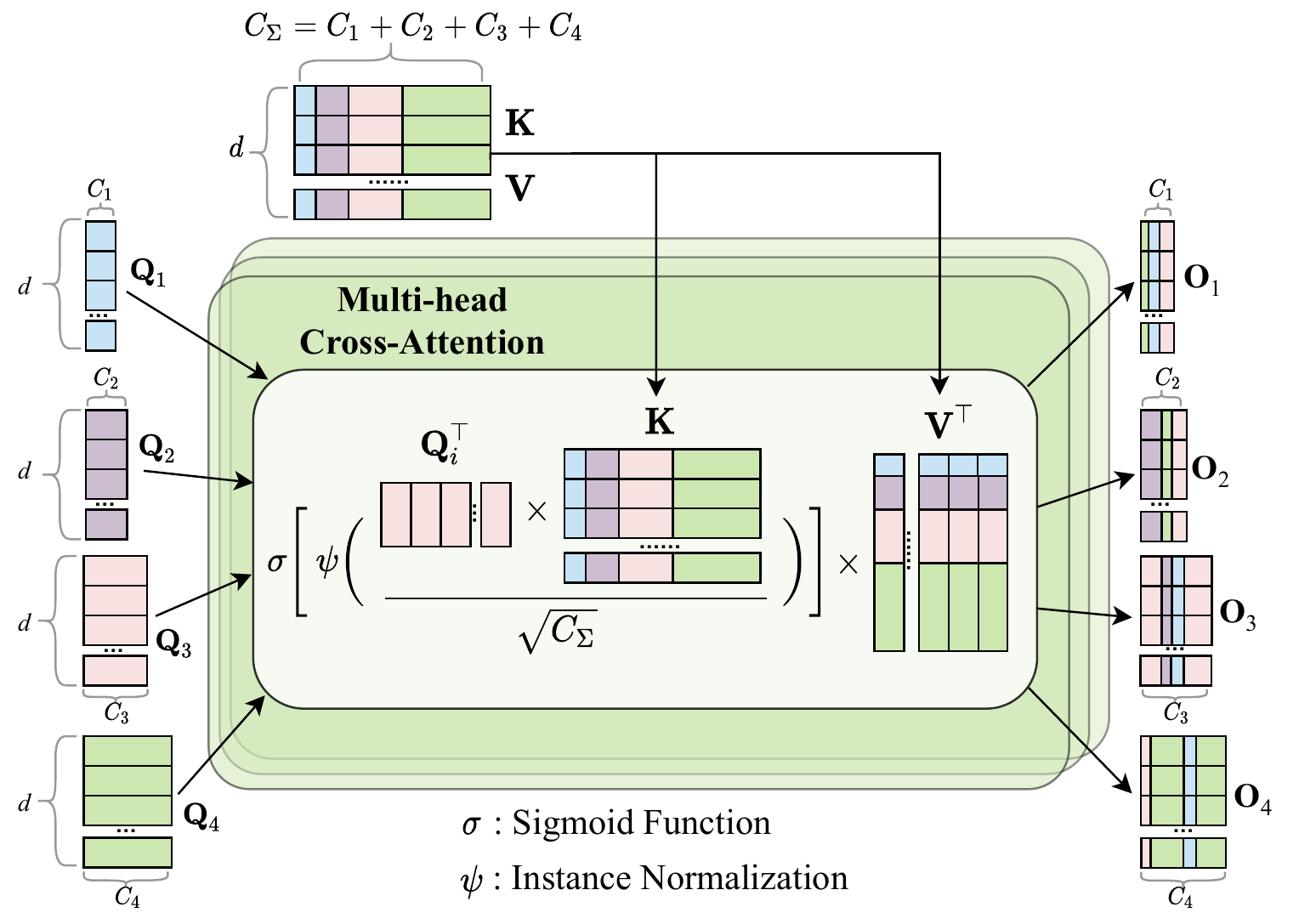} 
		\caption{Multi-head Cross-Attention.}
		\label{MHCA}
	\end{figure}

	\subsubsection{Multi-head Cross-Attention}
	
	The tokens are then fed into the multi-head channel cross-attention module, followed by a Multi-Layer Perceptron (MLP) with residual structure, to encode channel and dependencies for refining features from each U-Net encoder level using multi-scale features.  
	
	As shown in Fig.~\ref{MHCA}, the proposed CCT module contains five inputs, including four tokens $\mathbf{T}_i$ as queries and a concatenated token $\mathbf{T}_{\Sigma}$ as key and value:
	\begin{equation}
		\mathbf{Q}_i= \mathbf{T}_i W_{\mathbf{Q}_{i}},\mathbf{K}=\mathbf{T}_{\Sigma}W_{\mathbf{K}},\mathbf{V}=\mathbf{T}_{\Sigma}W_{\mathbf{V}}
	\end{equation}
	where $W_{\mathbf{Q}_{i}} \in \mathbb{R}^{C_i \times d}, W_\mathbf{K} \in \mathbb{R}^{C_\Sigma \times d}, W_\mathbf{V} \in \mathbb{R}^{C_{\Sigma} \times d}$ are weights of different inputs, $d$ is the sequence length (patch numbers) and $C_i (i=1,2,3,4)$ are the channel dimensions of the four skip connection layers. In our implementation $C_1=64,C_2=128,C_3=256,C_4=512$.

	With $\mathbf{Q}_{i} \in \mathbb{R}^{C_i \times d}, \mathbf{K} \in \mathbb{R}^{C_\Sigma \times d}, \mathbf{V} \in \mathbb{R}^{C_{\Sigma} \times d}$, the similarity matrix $\mathbf{M}_i$ are produced and the value $\mathbf{V}$ is weighted by $\mathbf{M}_i$ through a cross-attention (CA) mechanism:
	\begin{equation}
		\begin{aligned}
			\mathrm{CA}_i  = \mathbf{M}_i \mathbf{V^\top}  &= 
			\sigma\left[\psi\left(\frac {\mathbf{Q}^\top_i\mathbf{K}} {\sqrt{C_\Sigma}}\right)\right] \mathbf{V}^\top \\ & = \sigma\left[\psi\left(\frac{W^\top_{\mathbf{Q}_{i}} \mathbf{T}_i^\top \mathbf{T}_{\Sigma} {W}_{\mathbf{K}} }{\sqrt{C_\Sigma}}  \right) \right]{W}_{\mathbf{V}}^\top \mathbf{T}_{\Sigma}^\top
		\end{aligned}
	\end{equation}
	where $\psi(\cdot)$ and $\sigma(\cdot)$  denote the instance normalization\cite{InstanceNormalizationMissing_2017} and the softmax function, respectively.
	
	The major difference from the original self-attention is that we conduct the attention operation along the channel-axis rather than the patch-axis (see Fig.~\ref{TransComparison}), and we employ the instance normalization which can normalize the similarity matrix for each instance on the similarity maps so that the gradient can be smoothly propagated.
	In a $N$-head attention situation, the output after multi-head cross-attention is calculated as follow:
	\begin{equation}
		\mathrm{MCA}_i =(\mathrm{CA}_{i}^1+ \mathrm{CA}_{i}^2+,\dots,+\mathrm{CA}_{i}^N) / N
	\end{equation}
	where $N$ is the number of heads. Hereinafter, applying a MLP and residual operator, the output is obtained as follows: 
	\begin{equation}
		\mathbf{O}_i = \mathrm{{MCA}}_i+ \mathrm{MLP}( \mathbf{Q}_i + \mathrm{MCA}_i)
	\end{equation}
	We omitted the layer normalization (LN) in the equation for simplicity. The operation in Eq. (4) is repeated $L$ times to build a $L$-layer Transformer. In our implementation, $N$ and $L$ are both set to $4$, based on series of experiments with 2, 4, 8 and 12 layers for CCT and we empirically find the one with 4 layers and 4heads can achieve the best performance
	on the three datasets.
	Finally, the four outputs of the $L$-th layer $\mathbf{O}_1$, $\mathbf{O}_2$, $\mathbf{O}_3$ and $\mathbf{O}_4$ are reconstructed though an up-sampling operation followed by a convolution layer and concatenated with the decoder features $\mathbf{D}_1$, $\mathbf{D}_2$, $\mathbf{D}_3$ and $\mathbf{D}_4$, respectively.

	\subsection{CCA: Channel-wise Cross Attention for Feature Fusion in Decoder}
	In order to better fuse features of inconsistent semantics between the Channel Transformer and U-Net decoder, we propose a channel-wise cross attention module, which can guide the channel and information filtration of the Transformer features and eliminate the ambiguity with the decoder features.

	Mathematically, we take the $i$-th level Transformer output $\mathbf{O_i}\in\mathbb{R}^{C \times H\times W}$ and $i$-th level decoder feature map $\mathbf{D_i}\in\mathbb{R}^{C \times H\times W}$ as the inputs of Channel-wise Cross Attention.
	spatial squeeze is performed by a global average pooling (GAP) layer, producing vector $\mathcal{G}(\mathbf{X}) \in\mathbb{R}^{C \times 1\times 1}$ with its $k^{th}$ channel $\mathcal{G}(\mathbf{X})=\frac 1 {H\times W} \sum_{i=1}^H\sum_{j=1}^W \mathbf{X}^k(i,j)$.
	We use this operation to embed the global spatial information and then generate the attention mask:  
	\begin{equation}
		\mathbf{M}_i=\mathbf{L}_1\cdot\mathcal{G}(\mathbf{O_i})+\mathbf{L}_2\cdot\mathcal{G}(\mathbf{D_i})
	\end{equation}
	where $\mathbf{L}_1\in\mathbb{R}^{C\times C}$ and $\mathbf{L}_2\in\mathbb{R}^{C \times C}$ and being weights of two Linear layers and the ReLU operator $\delta (\cdot)$. 
	This operation in Eq. (5) encodes the channel-wise dependencies. Followed ECA-Net \cite{ECANetEfficientChannel_2020} which empirically showed avoiding dimensionality reduction is important for learning channel attention, we use a single Linear layer and sigmoid function to build the channel attention map. The resultant vector is used to recalibrate or excite $\mathbf{O}_i$ to
	$\mathbf{\hat{O}}_i=\sigma(\mathbf{M}_i)\cdot \mathbf{O_i}$, where the activation $\sigma(\mathbf{M}_i)$ indicates the importance of channels. 
	Finally, the masked $\mathbf{\hat{O}}_i$ is concatenated with the up-sampled features of the $i$-th level decoder.

	\begin{table}[t]
		\scriptsize
		\centering
		
		\begin{tabular}{@{} l @{\ \ } c @{\ \ } c @{\ \ } l @{\ \ } l @{\ \ } l @{\ \ \ } l@{}}
			\toprule
			\multirow{2}{*}{Method} & \multirow{2}{*}{Param} & \multirow{2}{*}{FlOPs}  & \multicolumn{2}{c}{GlaS} & \multicolumn{2}{c}{MoNuSeg} \\ \cmidrule(l){4-7} 
			& (M)  & (G)     &Dice (\%)            & IoU (\%)           & Dice (\%)          & IoU (\%) \\ \midrule
			U-Net      &14.8  &50.3     &85.45$\pm$1.25  &74.78$\pm$1.67  &76.45$\pm$2.62  &62.86$\pm$3.00 \\
			UNet++     &74.5  &94.6     &87.56$\pm$1.17  &79.13$\pm$1.70  &77.01$\pm$2.10  &63.04$\pm$2.54 \\
			AttUNet    &34.9  &101.9    &88.80$\pm$1.07  &80.69$\pm$1.66	&76.67$\pm$1.06	 &63.47$\pm$1.16 \\
			MRUNet     &57.2  &78.4     &88.73$\pm$1.17  &80.89$\pm$1.67	&78.22$\pm$2.47	 &64.83$\pm$2.87\\ 
			TransUNet  &105 &56.7    &88.40$\pm$0.74  &80.40$\pm$1.04	&78.53$\pm$1.06	 &65.05$\pm$1.28 \\
			MedT       &98.3  &131.5    &85.92$\pm$2.93  &75.47$\pm$3.46	&77.46$\pm$2.38	 &63.37$\pm$3.11 \\
			Swin-Unet  &82.3  &67.3    &89.58$\pm$0.57  &82.06$\pm$0.73	&77.69$\pm$0.94	 &63.77$\pm$1.15 \\
			\textbf{Ours} &65.6 &63.2   &\textbf{90.18$\pm$0.71$^\ast$}  &\textbf{82.96$\pm$1.06$^\ast$}	&\textbf{79.08$\pm$0.67}	 &\textbf{65.50$\pm$0.91}
			\\
			\bottomrule
		\end{tabular}
		\caption{ The three times 5-fold cross validation results on GlaS and MoNuSeg datasets. The Dice and IoU are in `mean$\pm$std' format. Symbol $\ast$ indicates that our method significantly outperformed others on that score (Student’s t-test at a level of 0.05 is used).}
		\label{SOTA1}
	\end{table}

	\begin{table}[t]
		\centering
		\footnotesize
		\begin{tabular}{@{}lcc@{}}
			\toprule
			Methods & Dice$\uparrow$        & HD$\downarrow$   \\ \midrule
			V-Net \shortcite{VNetFullyConvolutional_2016}                    & 68.81        & -           \\
			DARR \shortcite{DomainAdaptiveRelational_2020}                     & 69.77        & -           \\
			U-Net \shortcite{UNetConvolutionalNetworks_2015}        & 71.77        & 53.04        \\
			R50-U-Net                & 74.68        & 36.87        \\
			R50-AttUNet \shortcite{AttentionUNetLearning_2018}             & 75.57        & 36.97        \\
			TransUNet \shortcite{TransUNetTransformersMake_2021}               & 77.48        & 31.69 \\
			Swin-Unet \shortcite{SwinUnetUnetlikePure_2021}               & \textbf{79.13}        & \textbf{21.55}\\ 
			UCTransNet (w/o CCA)      & 78.99        & 30.29        \\
			UCTransNet-pre                & 75.54        & 38.97       \\
			\textbf{UCTransNet}                & 78.23        & 26.75        \\
			\bottomrule
		\end{tabular}
		\caption{Comparison with state-of-the-art segmentation methods on Synapse dataset. For simplicity, `R50-U-Net' and `R50-AttUNet' denote U-Net and Attention U-Net with ResNet-50 as backbone, respectively.}
		\label{SOTA_synapse}
	\end{table}
	
	\section{Experiments}
	
	\subsection{Datasets}
	
	We use Gland segmentation \cite{GlandSegmentationColon_2016}, MoNuSeg \cite{DatasetTechniqueGeneralized_2017,MultiOrganNucleusSegmentation_2020} and Synapse multi-organ segmentation dataset \cite{SegmentationOutsideCranial_2015} to evaluate our method. Gland segmentation dataset (GlaS) has 85 images for training and 80 for testing. MoNuSeg dataset has 30 images for training and 14 for testing. 
	Synapse has 30 abdominal CT scans in 8 abdominal organs (aorta, gallbladder, spleen, left kidney, right kidney, liver, pancreas, spleen, stomach), with 3779 axial CT images in total. Following \cite{TransUNetTransformersMake_2021}, we use a random split of 18 training cases (2212 axial slices) and 12 cases for validation.

	\begin{figure*}[!h]
		\centering
		\includegraphics[width=0.95\textwidth]{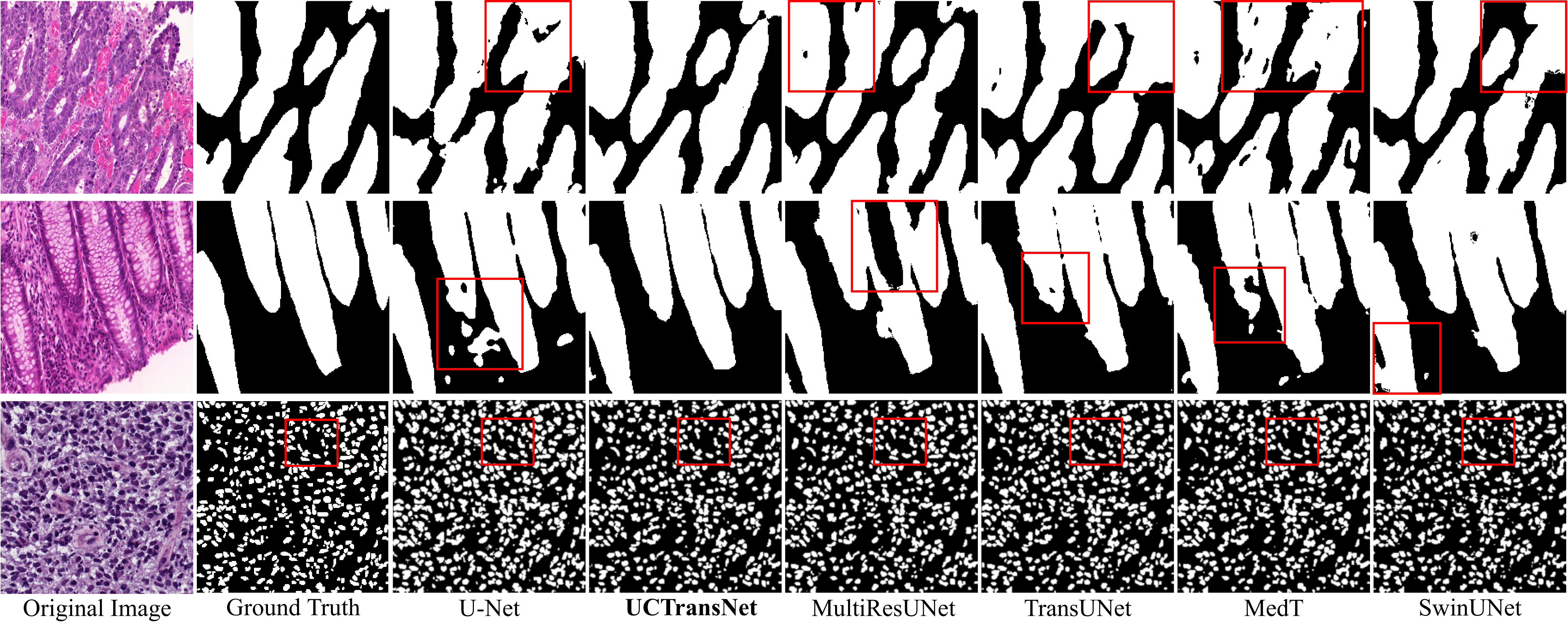} 
		\caption{The qualitative comparison on the GlaS and MoNuSeg datasets.}
		\label{SOTA_G_M}
	\end{figure*}
	
	\begin{figure}[!h]
		\centering
		\includegraphics[width=1.0\columnwidth]{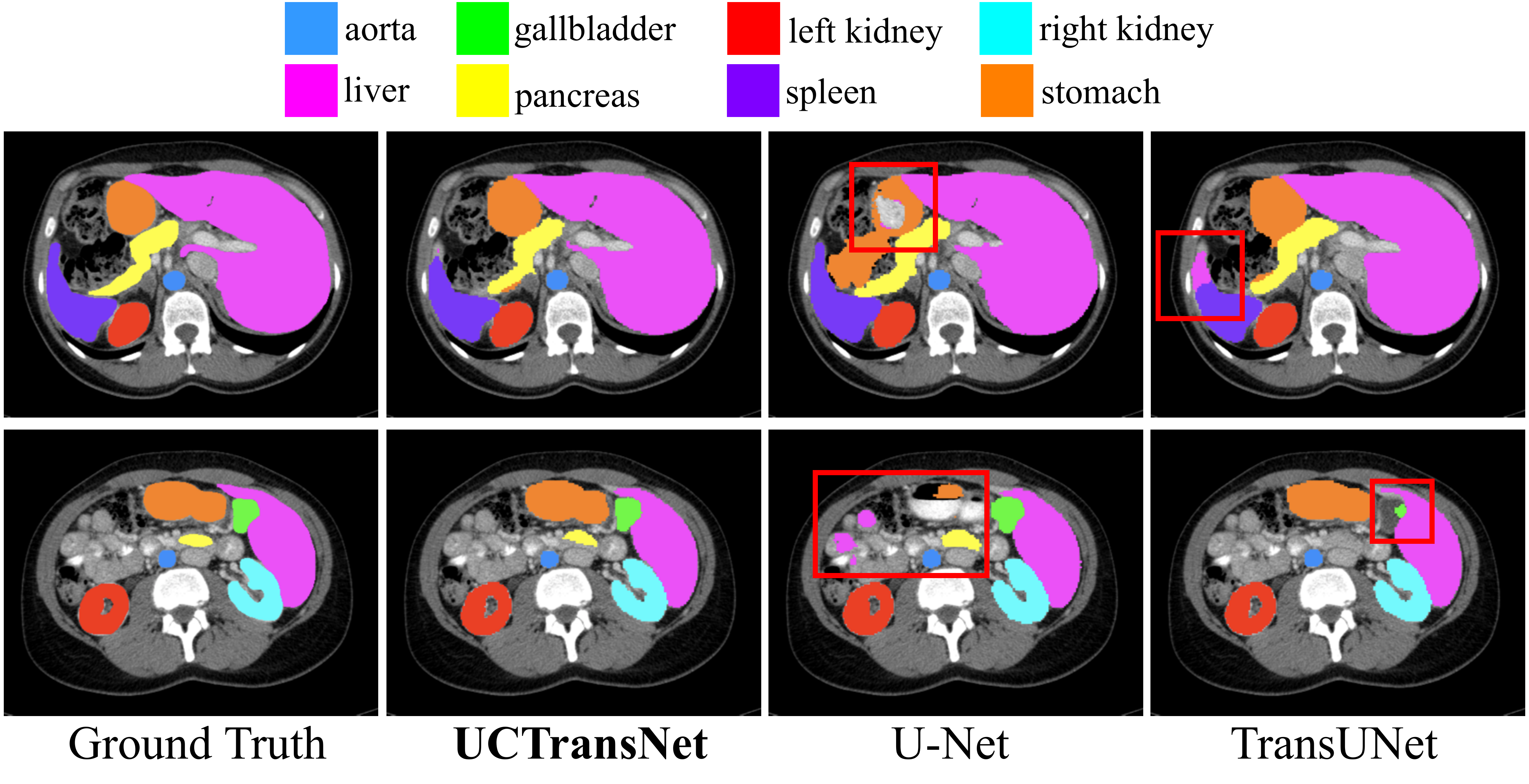} 
		\caption{The qualitative comparison on the Synapse dataset.}
		\label{SOTA_S}
	\end{figure}
	
	\subsection{Implementation Details}
	We implemented our model with PyTorch on a single NVIDIA A40 GPU card with 48 GB memory. To avoid over-fitting, we also performed two kinds of online data augmentations, including horizontal flipping, vertical flipping and random rotating. 
	We do not use any pre-trained weights to train the proposed UCTransNet.
	For GlaS and MoNuSeg we set the batch size to 4 following \cite{MedicalTransformerGated_2021}, and for Synapse we set it to 24 following \cite{TransUNetTransformersMake_2021}.
	The input resolution and patch size $P$ are set as $224\times224$ and $16$ for all the three datasets.
	We employ the Adam optimizer to train our model, where the initial learning rate is set to 0.001.
	We also employ the combined cross entropy loss and dice loss as our loss function to train our network.
	To make the results on the small datasets more convincing, we conduct a three times 5-fold cross validation (totally 15 CV), and obtain the mean result and std. 
	A statistical test is used to indicate our method significantly outperforms the comparable methods.
	For GlaS and MoNuSeg datesets we use dice coefficient (Dice) and Intersection over Union (IoU) as the evaluation metrics while for Synapse we report the Dice and Hausdorff Distance (HD). 
	Note that we use the same settings and loss function for training all the baselines.

	\subsection{Comparison with State-of-the-art Methods}
	To demonstrate the overall segmentation performance of the proposed UCTransNet, we compare it with other state-of-the-art methods.
	We compare UCTransNet with two types of methods for comprehensive evaluation, covering three UNet based method: UNet++, Attention U-Net, MultiResUNet and three state-of-the-art transformer based segmentation methods, including TransUNet, MedT, and Swin-Unet. 
	To make a fair comparison, their originally released codes and published settings are used in the experiment.
	We also introduce two strategies to optimize the models of UCTransNet.
	1) \textbf{Jointly training}: We optimize the convolution and CTrans parameters in U-Net and the two channel-wise cross attention parameters  together with a single loss;  2) \textbf{Pre-training}. We first train a U-Net, then the parameters in UCTransNet are further trained with the same data.

	Experimental results are reported in Table~\ref{SOTA1} where the best results are boldfaced.
	Table~\ref{SOTA1} shows that our method has consistent improvements over prior arts.
	In Table~\ref{SOTA_synapse}, similar observations and conclusions can be made, which once again validates that UCTransNet outperforms all others. 
	Additionally, the pre-training scheme not only achieves a faster convergence speed, but also obtains a better performance than the competing methods, even outperforms the jointly learning scheme on the MoNuSeg dataset. These observations suggest that the two proposed modules can be incorporated into the pre-trained U-Net model for improved segmentation performance.
	We also provide the parameter number and GFLOPs which show that our model achieves a good trade-off between effectiveness and efficiency.
	
	We visualize the segmentation results of the comparable models in Fig.~\ref{SOTA_G_M} and Fig.~\ref{SOTA_S}.
	The red boxes highlight regions where UCTransNet performs better than the other methods. It shows that our UCTransNet generates better segmentation results, which are more similar to the ground truth than the results of the baseline model. It can be easily seen that our proposed method not only highlights the right salient regions eliminating the confusing false positive lesions but also produces coherent boundaries. 
	These observations suggest that UCTransNet is capable of finer segmentation while preserving detailed shape information.

	\subsection{Ablation Studies}
	
	\subsubsection{Ablation Studies on Proposed Modules}
	As shown in Table~\ref{ablation}, `Base+CCT+CCA' is generally better than the other baselines on all datasets, which indicates the effectiveness of combination of the two modules.
	Our results shed new light on the importance of multi-scale multi-channel feature fusion in encoder-decoder framework for improving segmentation performance.
	
	\begin{table}[]
		\centering
		\footnotesize
		\begin{tabular}{@{}lcccc@{}}
			\toprule
			\small
			\multirow{2}{*}{Method} & \multicolumn{2}{c}{GlaS} & \multicolumn{2}{c}{MoNuSeg} \\ \cmidrule(l){2-5} 
			& Dice(\%)     & IoU(\%)    & Dice(\%)      & IoU(\%)     \\ \midrule
			Baseline (U-Net)             & 85.45        & 74.78      & 76.45         & 62.86       \\
			Baseline+CCT                 & 89.09        & 80.78      & 79.31         & 65.97       \\
			Baseline+CCA                & 87.09        & 78.10      & 76.84         & 63.85       \\
			Baseline+CCT+CCA             & \textbf{89.84}        & \textbf{82.24}      & \textbf{79.87}         & \textbf{66.68}     \\ \bottomrule
		\end{tabular}
		\caption{Ablation experiments on GlaS and MoNuSeg datasets. `CCT' denotes the proposed Channel Transformer and `CCA' denotes Channel-wise Cross Attention. The best results are boldfaced.}
		\label{ablation}
	\end{table}

	\subsubsection{Ablation Studies on the Number of Queries and Keys}
	\begin{figure}[!t]
		\centering
		\includegraphics[width=0.9\columnwidth]{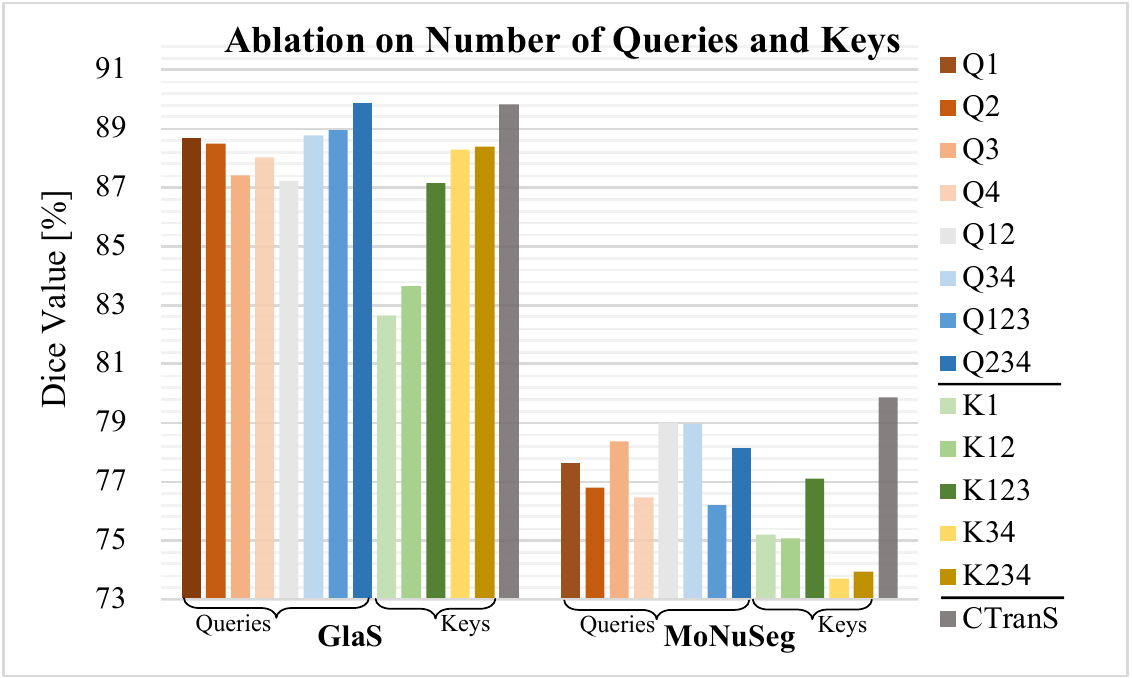} 
		\caption{Ablation study of the number of queries and keys on GlaS dataset and MoNuSeg dataset.}
		\label{Qnum}
	\end{figure}
	The previous experiments demonstrate that the CCT module in our model is effective for enhancing the skip connections. 
	In the  previous experiments,  the multi-scale features from all encoder levels engage into the CCT module, thus the number of queries is $4$ and the key is the concatenated representation consisting of the four scale features.
	
	We perform a series of experiments with respect to the amount of the skip connections between encoders and decoders as illustrated in Fig.~\ref{Qnum}. Note that the key vector is fixed, which is still consisting of the four scale features.
	We observe consistent improvements  with the increase of the number of skip connections. 
	The observation implies the usefulness of multi-scale features learned by different encoder levels, which validates our motivation. 
	It is interesting that `Q234' is slightly better than our model with all skip connections.
	Besides, we also keep the number of queries fixed and vary the key to verify  concatenating multi-scale features.
	From Fig.~\ref{Qnum}, it can be found that the performance improves with the scale of feature increasing until four scales, which demonstrates that more channels help capture accurate node features, which implies that transforming more scales of features to queries is better. 
	
	
	\begin{figure}[!t]
		\centering
		\includegraphics[width=0.9\columnwidth]{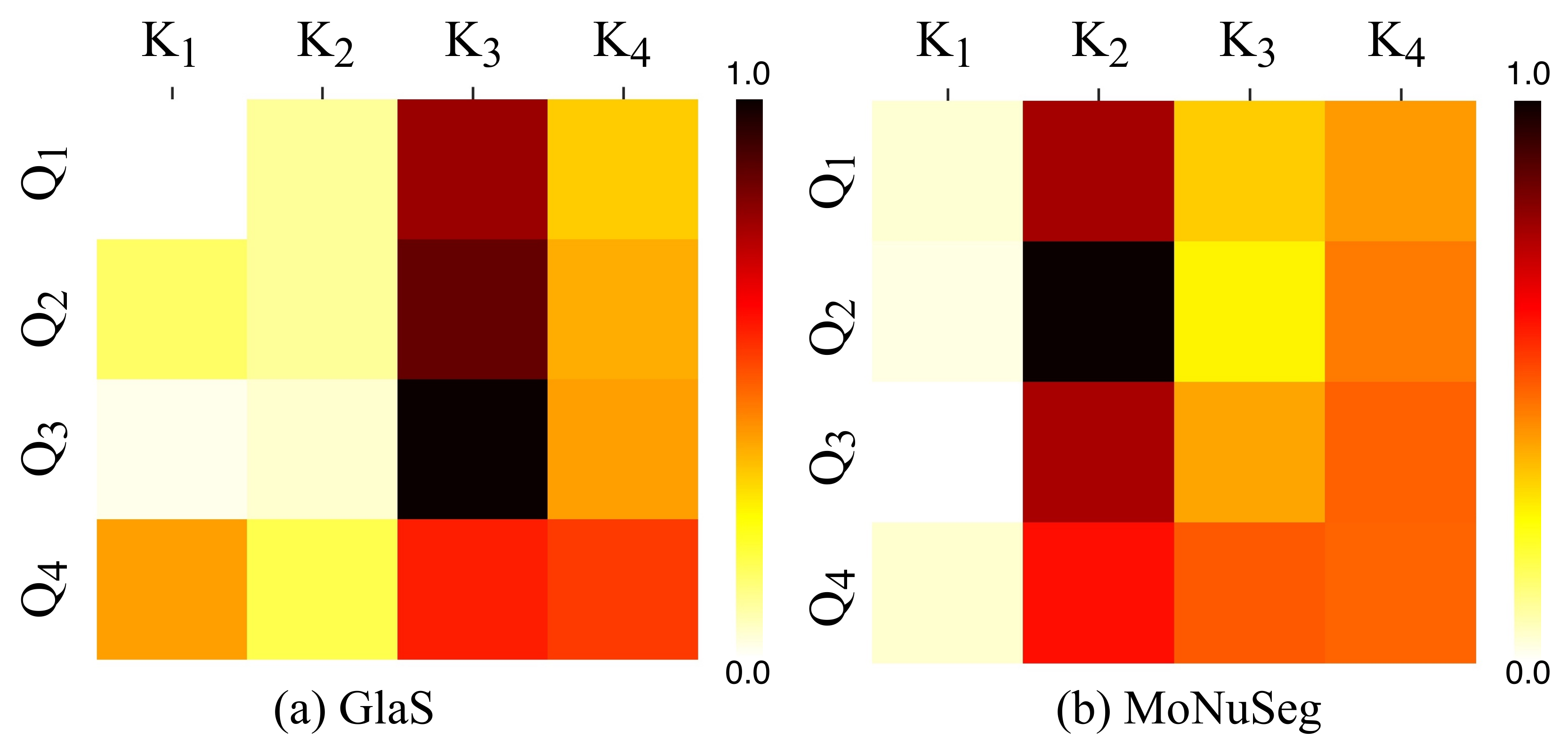} 
		\caption{Similarity Matrix of GlaS dataset (a) and MoNuSeg dataset (b). `$\mathbf{K_1}$' denotes the same feature as `$\mathbf{Q_1}$' concatenated in key.}
		\label{AttMap}
	\end{figure}

	\subsubsection{The Cross Attention Matrix in CCT Module}
	To perform a thorough evaluation of our UCTransNet, we visualize the cross attention distributions in our CCT module in Fig.~\ref{AttMap}.
	It is also interesting to investigate which encoder level has more confident correlation and is more important for segmentation.
	It can be seen that the `$\mathbf{K_2}$' and `$\mathbf{K_3}$' have a more confident correlation with the other encoder level on the GlaS and MoNuSeg dataset, respectively.
	The findings are consistent with the skip connection analysis in U-Net in Fig.~\ref{skip}. It explains why `L3' and `L2' achieve better performances on GlaS and MoNuSeg dataset, respectively.
	The fact implies the necessity to develop a multi-scale feature fusion to tackle the semantic gap problems, which also validates our motivation to build a global multi-scale channel-wise feature fusion model for effectively capturing the non-local semantic dependencies.

	\section{Conclusion}
	Accurate and automatic segmentation of medical images is a crucial step for clinical diagnosis and analysis.
	In this work, we introduced a Channel Transformer Segmentation network (UCTransNet) from the channel-wise perspective to provide precise and reliable automatic segmentation of medical images.
	By combining the strengths of multi-scale Channel-wise Cross fusion Transformer (CCT) and recurrent neural networks and Channel-wise Cross-Attention (CCA) in an end-to-end manner, the proposed approach significantly improves the state-of-the-art results in medical image segmentation on multiple benchmark datasets.
	With in-depth analysis and empirical evidence, we show the advantages of the UCTransNet model. It indeed successfully narrows the semantic gap and takes full advantage of the multi-scale features in the encoding stage.

	\section{Acknowledgements}
	This research was supported by the National Natural Science Foundation of China (No.62076059), the Fundamental Research Funds for the Central Universities (No. N2016001) and the Science Project of Liaoning province (2021-MS-105).

	\bibliography{CTUNet.bib}

\end{document}